\documentclass[10pt,twocolumn,letterpaper]{article}

\usepackage{iccv}
\usepackage{times}
\usepackage{epsfig}
\usepackage{graphicx}
\usepackage{amsmath}
\usepackage{amssymb}
\usepackage{color}
\usepackage{mathtools}
\usepackage{fixltx2e}
\usepackage{multirow}

\usepackage[pagebackref=true,breaklinks=true,letterpaper=true,colorlinks,bookmarks=false]{hyperref}

\iccvfinalcopy

\ificcvfinal\fi

\begin{document}


\title{Squared Earth Mover's Distance-based Loss for Training Deep Neural Networks}

\author{Le Hou\\
Computer Science Department\\
Stony Brook University\\
{\tt\small lehhou@cs.stonybrook.edu}
\and
Chen-Ping Yu\\
Psychology Department\\
Harvard University \\
{\tt\small chenpingyu@fas.harvard.edu}
\and
Dimitris Samaras\\
Computer Science Department\\
Stony Brook University\\
{\tt\small samaras@cs.stonybrook.edu}
}

\maketitle

\begin{abstract}
	In the context of single-label classification, despite the huge success of deep learning, the commonly used cross-entropy loss function ignores the intricate inter-class relationships that often exist in real-life tasks such as age classification. In this work, we propose to leverage these relationships between classes by training deep nets with the exact squared Earth Mover's Distance (also known as Wasserstein distance) for single-label classification. The EMD$^2$ loss uses the predicted probabilities of all classes and penalizes the miss-predictions according to a ground distance matrix that quantifies the dissimilarities between classes. We demonstrate that on datasets with strong inter-class relationships such as an ordering between classes, our exact EMD$^2$ losses yield new state-of-the-art results. Furthermore, we propose a method to automatically learn this matrix using the CNN's own features during training. We show that our method can learn a ground distance matrix efficiently with no inter-class relationship priors and yield the same performance gain. Finally, we show that our method can be generalized to applications that lack strong inter-class relationships and still maintain state-of-the-art performance. Therefore, with limited computational overhead, one can always deploy the proposed loss function on any dataset over the conventional cross-entropy.
\end{abstract}

\section{Introduction}

Deep neural networks (DNNs) have become the preferred method for most machine learning applications, due to their ability to automatically learn optimal features and classifiers from inputs in an end-to-end fashion. In addition to superior performance as compared to conventional approaches, another reason for the popularity is their wide-range of applicability which includes convolutional neural networks (CNNs) for computer vision \cite{he2015deep,simonyan2014very}, recurrent neural networks (RNNs) for natural language processing \cite{socher2011parsing,liu2014recursive}, hybrid networks that combine CNN and RNN layers for speech recognition and audio processing \cite{hannun2014deep,deng2014ensemble}, and more. In general, most DNNs are trained under one of two tasks: regression and classification. In a regression task, the network learns to generate a real-valued output that matches the ground-truth \cite{belagiannis2015robust,dong2014learning}. In a classification task, the network learns to categorize an input to one of the training classes \cite{eidinger2014age,zagoruyko2016wide,he2015deep,simonyan2014very,li2015convolutional}. Other tasks such as detection and segmentation are often cast as classification tasks using a sliding window or a multi-label classification approach.

\begin{figure}[t]
\begin{center}
   \includegraphics[trim=0cm 0cm 0cm 0cm,clip, width=0.98\linewidth]{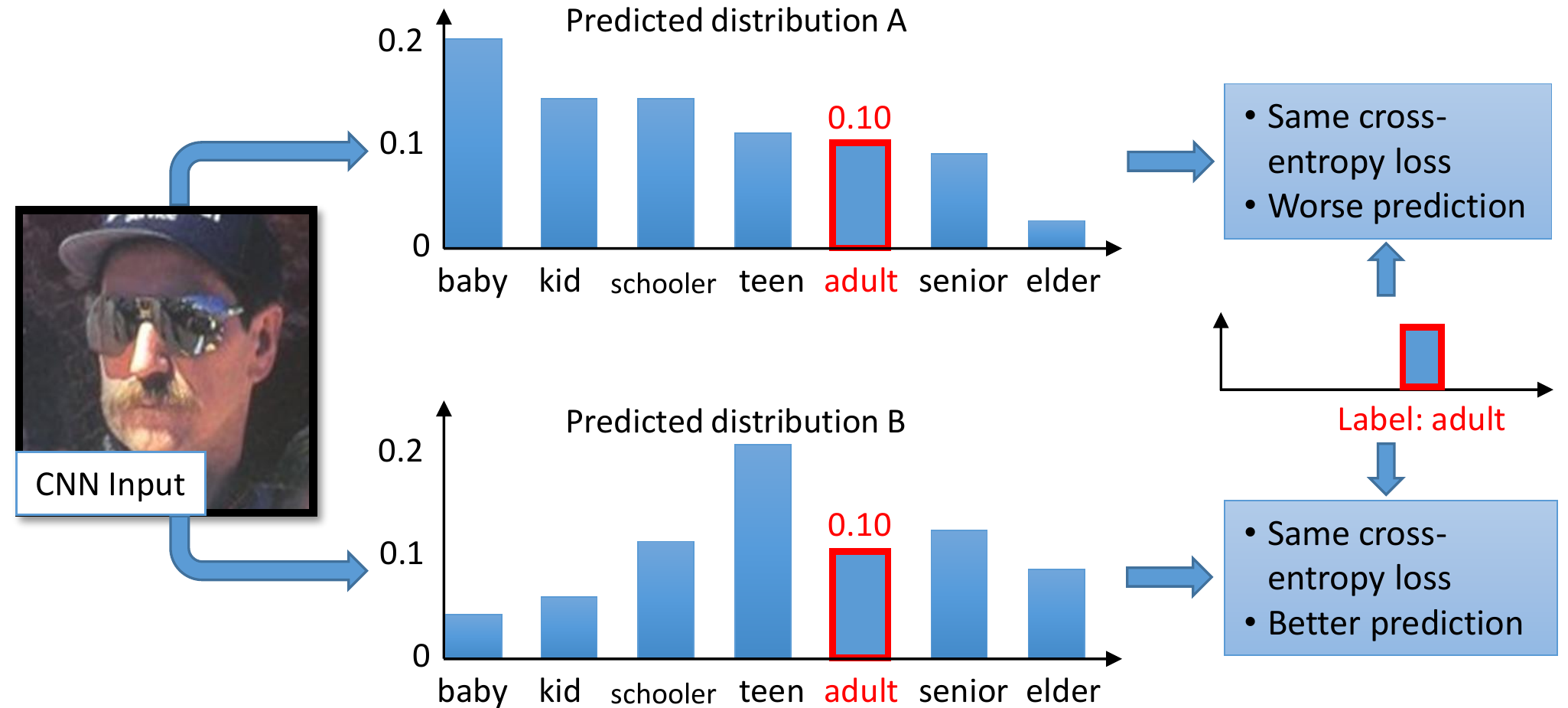}
\end{center}
   \caption{In many classification tasks, there are relationships or even orderings between classes. However the cross-entropy loss ignores these relationships and only focuses on the predicted probability of the ground truth class. In this example, the two given predicted distributions have the same cross-entropy loss. But clearly predicted distribution B is preferable to A.}
\label{fig:first_fig}
\end{figure}


To train a multi-class single-label classification network, softmax cross-entropy loss is by far the most popular loss function for the training regime, where the ground-truth is a binary vector consisting of a value 1 at the correct class index, and 0s everywhere else \cite{lecun1998gradient,krizhevsky2012imagenet}. During training, the objective is to minimize the negative log-likelihood of the loss by multiplying the network's predictions to the binary ground-truth vectors. While the softmax cross-entropy loss has been successfully used across all applicable fields, the loss function does not take into account inter-class relationships which can be very informative. For example (Fig. \ref{fig:first_fig}), we want to estimate human age-groups from face images. Given an image of an adult, if network A's output probability of the adult class is 0.1 with a max probability of 0.2 at the baby class, and network B's output probability of the adult class is 0.1 with a max probability of 0.2 at the teen class, both networks would achieve the same softmax cross-entropy loss, while network B's output probabilities are clearly closer to the ground-truth.


In this work, we show how the exact squared Earth Mover's Distance (EMD) \cite{rubner2000earth} can be applied both as a stand-alone loss function or as a regularization term for multi-class classification problems using CNNs. The EMD is also known as the Wasserstein distance \cite{arjovsky2017wasserstein,bogachev2012monge}, which is the minimal cost required to transform one distribution to another \cite{rubner2000earth}. Recent work formulated an approximate Wasserstein loss for supervised multi-class multi-label learning using a linear model, and applied it to classification problems with predefined inter-class similarity metrics \cite{frogner2015learning,martinez2016relaxed}. In contrast, we show that an \textit{exact} (without approximation) squared EMD (EMD$^2$)-based loss exists for training single-label deep learning models directly, either with known inter-class relationship or without any priors on the inter-class relationships. We choose to use EMD$^2$ instead of EMD as the loss function because squaring usually leads to faster convergence with gradient descent \cite{shalev2011stochastic,luenberger1973introduction}. Our experiments show that CNNs trained with our EMD$^2$ loss result in better performance than CNNs with the standard softmax cross-entropy loss, and achieve state-of-the-art results on multiple datasets. 

Computing the EMD requires a predefined ground distance matrix that quantifies the dissimilarities between classes. Existing work assumes a ground distance matrix. For example, when classes are ordered, the ground distance matrix has one dimensional embedding. Therefore, a closed-form solution exists for computing the EMD \cite{levina2001earth}. Otherwise, the ground distance matrix was obtained externally \cite{frogner2015learning}. These approaches are limited by the need for reasonable assumptions about the ground distance matrix.

In this work, we also show how to learn the ground distance matrix using the CNN's own features during training with limited additional computational cost, obviating the need for initial assumptions about the inter-class relationships. We propose to use EMD$^2$ computed using the estimated ground distance matrix as a regularization term in CNN training. We call this self-guided training with EMD$^2$-based regularization. We verify the learned matrix on datasets with known ordered-classes. Examples of such datasets include age estimation \cite{eidinger2014age,escalera2015chalearn,gallagher2009understanding,rothe2016deep,dong2016automatic}, image aesthetics \cite{kong2016photo}, facial attractiveness prediction \cite{sutic2010automatic} and others \cite{ma2016texture}. Experiments show that our self-guided CNN with EMD$^2$-based regularization performs as well as CNNs with EMD$^2$-based loss computed using ground distance matrices based on prior knowledge. Furthermore, we show that on datasets with weak inter-class relationships, the learned ground distance matrix does not capture spurious inter-class relationships that could adversely affect performance.


We claim three major contributions in this paper:
\begin{enumerate}
\setlength\itemsep{-0.2em}
\item For the first time, we show how an \textit{exact} EMD$^2$-based loss function can be used to train CNNs.
\item For the first time, we propose a method to efficiently discover inter-class relationships during training and use the discovered inter-class relationships as a ground distance matrix for CNN training with an \textit{exact} EMD$^2$-based regularization.
\item We improve state-of-the-art performance on datasets with strong inter-class relationships and avoid adverse effects on datasets with weak inter-class relationships.
\end{enumerate}


\section{EMD$^2$-based Loss}
\label{sec:method}
In this section, we first introduce the standard softmax cross-entropy loss and discuss its drawbacks in detail. Then, we formulate the EMD$^2$ and show it can be used as a loss function for classification problems with known assumptions on inter-class relationships.

\subsection{Softmax cross-entropy loss}

The softmax cross-entropy loss is a log-likelihood based loss function, that combines a softmax layer with a cross-entropy loss function. For a single-label classification problem with $C$ classes, a network's softmax layer outputs a probability distribution $\mathbf{p}$ of length $C$, with its $i$-th entry $\mathbf{p}_i$ being the predicted probability of the $i$-th class. The soft-max guarantees that $\sum_i \mathbf{p}_i = 1$. We denote the ground truth as a binary vector $\mathbf{t}$ of length $C$. Also $\sum_i \mathbf{t}_i = 1$. Given a training example, the cross-entropy loss between prediction $\mathbf{p}$ and ground truth vector $\mathbf{t}$ is defined as $\mathrm{E_X}(\mathbf{p}, \mathbf{t}) = -\sum_{i=1}^C \big(\mathbf{t}_i \mathrm{log}(\mathbf{p}_i)\big)$. We assume that the $k$-th class is the ground truth label: $\mathbf{t}_k = 1$ and $\mathbf{t}_i = 0$ for $i\ne k$. Thus the differentiation of $\mathrm{E_X}(\mathbf{p}, \mathbf{t})$ is:
\begin{equation}
\label{eq:soft-max-dev}
\mathrm{E}_X'(\mathbf{p}, \mathbf{t}) = -\mathbf{p}_k' / \mathbf{p}_k \text{.}
\end{equation}
It is obvious to see that the backpropagation of a DNN with cross-entropy loss only depends on $\mathbf{p}_k$. This is less robust compared to a loss function that depends on all entries of $\mathbf{p}$ as argued in Fig. \ref{fig:first_fig}

\subsection{EMD$^2$-based loss on ordered-classes}

Here, we first define the Earth Mover's Distance (EMD), and explain how an EMD$^2$-based loss function models inter-class relationships. Then, we define the problem of ordered-class classification and show when the exact EMD$^2$ function can be computed by a closed-form equation.

\subsubsection{Earth Mover's Distance}

We assume that a well performing CNN should predict class distributions such that classes closer to the ground truth class should have higher predicted probabilities than classes that are further away. We formulate this using the Earth Mover's Distance (EMD). The EMD is defined as the minimum cost to transport the mass of one distribution (histogram) to the other.

Mass transportation defines the problem of transporting mass from a set of supplier clusters to a set of consumer clusters. Its formal definition \cite{rubner2000earth} is: Let $\mathbf{p} = \{(\mathbf{a}_1, \mathbf{p}_1), (\mathbf{a}_2, \mathbf{p}_2), \dots, (\mathbf{a}_C, \mathbf{p}_C)\}$ be the supplier signature (distribution or histogram) with $C$ clusters (bins), where $\mathbf{a}_i$ represents each cluster and $\mathbf{p}_i$ is the mass (value) in each cluster. Let $\mathbf{t} = \{(\mathbf{b}_1, \mathbf{t}_1), (\mathbf{b}_2, \mathbf{t}_2), \dots, (\mathbf{b}_{C'}, \mathbf{t}_{C'})\}$ be the consumer signature. Let $\mathbf{D}$ be the ground distance matrix where its $i,j$-th entry $\mathbf{D}_{i,j}$ is the distance between $\mathbf{a}_i$ and $\mathbf{b}_j$. Matrix $\mathbf{D}$ is usually defined as the $l$-norm distance between clusters:
\begin{equation}
\mathbf{D}_{i,j} = ||\mathbf{a}_i - \mathbf{b}_j||_l
\label{eq:l-norm}
\end{equation}
Let $\mathbf{F}$ be the transportation matrix where its $i,j$-th entry $\mathbf{F}_{i,j}$ indicates the mass transported from $\mathbf{a}_i$ to $\mathbf{b}_j$. A valid transportation satisfies four constraints. First, the amount of mass transported must be positive. Second, the amount of mass transported from a supplier cluster $\mathbf{p}_i$ must not exceed its total mass. Third, the amount of mass transported to a consumer cluster $\mathbf{t}_j$ must not exceed its total mass. Finally, the total flow must not exceed the total mass that can be transported. These four conditions can be summarized respectively below:
\begin{equation}
\mathbf{F}_{i,j} \ge 0 \quad \text{for all $i$, $j$.}
\label{eq:constraint1}
\end{equation}
\begin{equation}
\sum\limits_{j=1}^{C'} \mathbf{F}_{i,j} \le \mathbf{p}_i \quad \text{for all $i$.}
\label{eq:constraint2}
\end{equation}
\begin{equation}
\sum\limits_{i=1}^{C} \mathbf{F}_{i,j} \le \mathbf{t}_j \quad \text{for all $j$.}
\label{eq:constraint3}
\end{equation}
\begin{equation}
\sum\limits_{i=1}^{C}\sum\limits_{j=1}^{C'} \mathbf{F}_{i,j} = \mathrm{min} \Big(\sum\limits_{i=1}^{C}\mathbf{p}_i, \sum\limits_{j=1}^{C'}\mathbf{t}_i \Big) \text{.}
\label{eq:constraint4}
\end{equation}
Under the constraints defined above, the overall cost of flow $\mathbf{F}$ is defined as:
\begin{equation}
\mathrm{W}(\mathbf{b}, \mathbf{t}, \mathbf{F}) = \sum\limits_{i=1}^{C}\sum\limits_{j=1}^{C'} \mathbf{D}_{i,j} \mathbf{F}_{i,j} \text{.}
\label{eq:work_cost}
\end{equation}

The EMD between two vectors, denoted as $\mathrm{EMD}(\mathbf{p}, \mathbf{t})$ is the minimum cost of work that satisfies the constraints in Eq. \ref{eq:constraint1}, \ref{eq:constraint2}, \ref{eq:constraint3}, \ref{eq:constraint4}, normalized by the total flow.
\begin{equation}
\mathrm{EMD}(\mathbf{p}, \mathbf{t}) = \inf\limits_{\mathbf{F}} \, \dfrac{\sum\limits_{i=1}^C\sum\limits_{j=1}^{C'} \mathbf{D}_{i,j} \mathbf{F}_{i,j}}{\sum\limits_{i=1}^C\sum\limits_{j=1}^{C'} \mathbf{F}_{i,j}} \text{.}
\label{eq:emd}
\end{equation}


\subsubsection{Ground distance matrix of ordered-classes}

Computing the EMD between two distributions requires a predefined matrix, the ground distance matrix $\mathbf{D}$ which is unknown in most cases. However, in classification tasks with ordered classes we can define $\mathbf{D}$. By ordered classes we mean classes that can be represented as real numbers, for example, human age ranges or aesthetic preference levels. The difference between ordered-class classification and regression is that in the problem of ordered-class classification, the ground truth labels and predictions are discrete. Hence, in practice better performance can be achieved using a multi-class classification model \cite{golik2013cross} instead of a regression model, and the ground distances between those classes can be based on their inherent ordering.

Without loss of generality, we assume that in all ordered-class classification problems, the classes are ranked as $\mathbf{t}_1, \mathbf{t}_2, \mathbf{t}_3, \dots, \mathbf{t}_C$ and the distance between $\mathbf{t}_i$ and $\mathbf{t}_j$ is $|i-j|$.


\subsubsection{EMD$^2$ loss for ordered-class classification}
\label{sec:emd_for_order}


EMD has been shown to be equivalent to Mallows distance which has a closed-form solution \cite{levina2001earth}, if the ground distance matrix $\mathbf{D}$ and distributions $\mathbf{p}$ and $\mathbf{t}$ satisfy certain conditions, as shown in \cite{levina2001earth}. We will show that these required conditions are satisfied in ordered-class classification problems.

The first condition is that the two distributions $\mathbf{p}$ and $\mathbf{t}$ to be compared must have equal mass: $\sum_i \mathbf{p}_i = \sum_j \mathbf{t}_j$. Note that this condition is always satisfied if $\mathbf{p}$ is produced by a softmax layer, as the output vector of a softmax layer is a normalized probability density function that sums to 1. And since the number of classes in the predicted distribution is the same as the target distribution, then $C=C'$.

The second condition is that the ground distance matrix $\mathbf{D}$ must have an one-dimensional embedding. Assuming $\mathbf{p}_1, \mathbf{p}_2, \dots, \mathbf{p}_C$ and $\mathbf{t}_1, \mathbf{t}_2, \dots, \mathbf{t}_{C'}$ are sorted according to their inherent rank values without loss of generality, this condition can be expresses as $\mathbf{D}_{i,j} = S(j-i)$, for a constant $S$ and all $i$, $j$ that $i\le j$. Clearly, this assumption can always be satisfied in ordered-class classification problems.

The third and final condition is that the distributions to be compared must be sorted vectors. This condition is also always satisfied since we assumed $\mathbf{p}_1, \mathbf{p}_2, \dots, \mathbf{p}_C$ and $\mathbf{t}_1, \mathbf{t}_2, \dots, \mathbf{t}_{C'}$ are sorted without loss of generality. Then, based on the conclusion by Levina \etal \cite{levina2001earth}, the normalized EMD can be computed exactly and in closed-form:
\begin{equation}
\mathrm{EMD} (\mathbf{p}, \mathbf{t}) = \Big(\frac{1}{C}\Big)^{\frac{1}{l}} ||\mathrm{CDF}(\mathbf{p}) - \mathrm{CDF}(\mathbf{t})||_l \text{,}
\end{equation}
where $\mathrm{CDF}(\cdot)$ is a function that returns the cumulative density function of its input.

We use $l=2$ for Euclidean distance and also for $\mathbf{D}$ in  Eq. \ref{eq:l-norm}. Dropping the normalization term, we obtain the final EMD$^2$ loss $\mathrm{E}_E$ as:
\begin{equation}
\label{eq:emd-loss-ordered}
\mathrm{E}_E (\mathbf{p}, \mathbf{t}) = \sum\limits_{i=1}^C \Big(\mathrm{CDF}_i(\mathbf{p}) - \mathrm{CDF}_i(\mathbf{t})\Big)^2 \text{,}
\end{equation} where $\mathrm{CDF}_i(\mathbf{p})$ is the $i$-th element of the CDF of $\mathbf{p}$. This equation is directly applicable to ordered-class classification problems. Note that we choose to use EMD$^2$ instead of EMD as the loss function because it usually converges faster and is easier to optimize with gradient descent \cite{shalev2011stochastic,luenberger1973introduction}.

\subsubsection{Derivative of the EMD$^2$ loss with ordered-classes}
To derive the derivatives of $\mathrm{E}_E$ with respect to the network parameters, we first rewrite $\mathrm{E}_E$ as: $\mathrm{E}_E(\mathbf{p}, \mathbf{t}) = (\mathbf{p}_1 - \mathbf{t}_1)^2 + (\mathbf{p}_1 + \mathbf{p}_2 - \mathbf{t}_1 - \mathbf{t}_2)^2 + \dots + \Big(\sum_i\mathbf{p}_i - \sum_j\mathbf{t}_j\Big)^2$. Each prediction $\mathbf{p}_i$ is a function of the network parameters. Differentiating $\mathrm{E}_E(\mathbf{p}, \mathbf{t})$ with respect to the network parameters yields:
\begin{equation}
\begin{split}
\label{eq:emd-dev-ordered}
& \mathrm{E}_E'(\mathbf{p}, \mathbf{t}) = 2\sum\limits_{n=1}^C \Bigg(\Big(\sum\limits_{i=1}^n\mathbf{p}_i - \sum\limits_{j=1}^n\mathbf{t}_j\Big) \Big(\sum\limits_{i=1}^n\mathbf{p}_i'\Big)\Bigg) \\
= & 2\mathbf{p}_1' \Big( \sum\limits_{i=1}^C \big(C-i+1\big)\big(\mathbf{p}_i - \mathbf{t}_i\big)\Big) \\
+ & 2\mathbf{p}_2' \Big( \sum\limits_{i=1}^C \big(C-i+1\big)\big(\mathbf{p}_i - \mathbf{t}_i\big) - \mathbf{p}_1 + \mathbf{t}_1 \Big) + \dots \\
+ & 2\mathbf{p}_C' \Big( \sum\limits_{i=1}^C \big(C-i+1\big)\big(\mathbf{p}_i - \mathbf{t}_i\big) - \sum\limits_{i=1}^{C-1}(C-i)\big(\mathbf{p}_i - \mathbf{t}_i\big) \Big)
\end{split}
\end{equation}
The coefficients of $\mathbf{p}_i'$ can be propagated using the standard backpropagation method.

Comparing Eq. \ref{eq:emd-dev-ordered} with Eq. \ref{eq:soft-max-dev}, we can see that the backpropagation of a network trained with cross-entropy loss is only based on $\mathbf{p}_k$ and $\mathbf{p}_k'$, whereas the backpropagation of a network trained with EMD$^2$ loss is based on all elements of $\mathbf{p}$ and $\mathbf{p}'$.

\section{Self-Guided EMD}
\label{sec:method_2}
Sec. \ref{sec:emd_for_order} showed the formulation of EMD$^2$ loss for ordered classes for which the ground distance matrix $\mathbf{D}$ can be easily assumed. However, in general the matrix $\mathbf{D}$ is unknown. In this section, we show how to compute a ground distance $\mathbf{D}$ using empirical evidence. Furthermore, we propose a method that computes the EMD$^2$-based loss between prediction $\mathbf{p}$ and ground truth $\mathbf{t}$ with $O(C)$ time complexity for single-label classification problems. Finally, for classification problems in general, we propose to use EMD$^2$ as a regularization term to the cross-entropy loss. Note that the proposed regularization term does not compete with other regularization terms such as $L2$ weight decay. One can apply all of these regularization terms for training.

\subsection{Estimating ground distances}
\label{sec:estimating_gd_matrix}
We focus on estimating the ground distance matrix $\mathbf{D}$ expressed in Eq. \ref{eq:l-norm}. Note that for classification problems, the predicted classes (supplier signatures) and the ground truth classes (consumer signatures) are the same set of classes. In other words $\mathbf{a}_1\vcentcolon=\mathbf{b}_1$, $\mathbf{a}_2\vcentcolon=\mathbf{b}_2$, $\dots$. We will use $\mathbf{a}_{1,2,\dots,C}$ to indicate the same set of classes in the future. We estimate $\mathbf{D}$ first by estimating all $\mathbf{a}_i$ (equivalently $\mathbf{b}_i$) directly. Then we compute an initial estimation of $\mathbf{D}$, denoted by $\mathbf{\bar D}$ directly from Eq. \ref{eq:l-norm}. Finally we postprocess $\mathbf{\bar D}$ to obtain an estimated $\mathbf{D}$.

To estimate each $\mathbf{a}_i$, we extract features on all instances of the $i$-th class, and use the centroid of these feature vectors as an initial estimate of $\mathbf{a}_i$, denoted as $\mathbf{\bar a}_i$. To extract the features of one instance, we follow a standard method \cite{girshick2014rich}: we use the second-to-last layer neural responses of the CNN that is being trained, as feature vectors. Note that we $L1$ normalize each instance's CNN features. Intuitively, because CNNs learn to \textit{linearly} separate classes with the second-to-last layer features (there is no subsequent non-linearity), it is meaningful to \textit{average} the feature vectors to compute the class centroids.


We denote $\mathbf{\bar D}$ as the initial estimation of the ground distance matrix where its $i,j$-th entry $\mathbf{\bar D}_{i,j} = ||\mathbf{\bar a}_i - \mathbf{\bar a}_j||_l$. We observe in practice that the CNN features cannot provide sufficient class separation before the network has partially converged. As a result, many entries of matrix $\mathbf{\bar D}$ are close to zero, indicating that class centroids are not well separated. To address this, we map each row of $\mathbf{\bar D}$ onto uniformly distributed values: each entry is mapped to its percentile value in its row. Formally, denoting the transformed matrix as $\mathbf{B}$, this operation is formulated as:
\begin{equation}
\mathbf{B}_{i,j} = \frac{1}{C} \, \mathrm{R}\big(\mathbf{\bar D}_{i,j}, \{\mathbf{\bar D}_{i,1}, \mathbf{\bar D}_{i,2}, \dots, \mathbf{\bar D}_{i,C}\}\big) \text{,}
\label{eq:percentile}
\end{equation} where $\mathrm{R}\big(\mathbf{\bar D}_{i,j}, \{\mathbf{\bar D}_{i,1}, \mathbf{\bar D}_{i,2}, \dots, \mathbf{\bar D}_{i,C}\}\big)$ returns the number of elements in the set $\{\mathbf{\bar D}_{i,1}, \mathbf{\bar D}_{i,2}, \dots, \mathbf{\bar D}_{i,C}\}$ that is smaller than $\mathbf{\bar D}_{i,j}$. After this transformation, all entries of $\mathbf{B}$ are between 0 and 1, and our final estimation of the ground distance matrix $\mathbf{D}$ is a symmetric matrix obtained below:
\begin{equation}
\label{eq:ground_distance_calculation}
\mathbf{D} = (\mathbf{B} + \mathbf{B}^T) / 2 \text{.}
\end{equation} Note that based on this definition, $\mathbf{D}_{i,i} = 0$ for all $i$.

\subsection{Self-guided EMD$^2$ regularization}

We now describe how to calculate the EMD between $\mathbf{p}$ and $\mathbf{t}$, defined by Eq. \ref{eq:emd}. In the case of single-label classification, the consumer's mass vector (target distribution) $\mathbf{t}$ is a binary vector where only the index of the ground truth class $k$ equals to 1: $\mathbf{t}_k = 1$. According to the constraint defined by Eq. \ref{eq:constraint3}, all mass must be transferred to the $k$-th cluster. Thus, the transportation matrix must satisfy $\mathbf{F}_{i,j} = 0$ if $j\ne k$, otherwise $\mathbf{F}_{i,j} = \mathbf{p}_i$. The resulting EMD is:
\begin{equation}
\label{eq:emd_orderless_classes}
\mathrm{EMD}(\mathbf{p}, \mathbf{t}) = \dfrac{\sum\limits_{i=1}^{C}\mathbf{p}_i\mathbf{D}_{i,k}}{\sum\limits_{i=1}^{C}\mathbf{p}_i} = \sum\limits_{i=1}^{C}\mathbf{p}_i\mathbf{D}_{i,k} \text{.}
\end{equation}
The computational complexity of EMD by Eq. \ref{eq:emd_orderless_classes} is $O(C)$.

The exact EMD defined by Eq. \ref{eq:emd_orderless_classes} can be directly used as a loss function. Its derivative with respect to network parameters is:
\begin{equation}
\label{eq:emd-dev-orderless}
\mathrm{EMD}'(\mathbf{p}, \mathbf{t}) = \sum\limits_{i=1}^{C}\mathbf{p}_i'\mathbf{D}_{i,k} \text{.}
\end{equation}
In practice, the optimization does not converge to a desired local optimum using Eq. \ref{eq:emd_orderless_classes} directly. We observed that using Eq. \ref{eq:emd-dev-orderless} for gradient descent ends up lowering the predicted probabilities of all classes, which leads the model to converge to a local minimum with uniformly distributed predictions. To address this optimization problem, we modify Eq. \ref{eq:emd_orderless_classes} and use it as a regularizer instead of a stand-alone loss function. Additionally, for faster optimization with gradient descent, we use $\mathbf{p}_i^2$ instead of $\mathbf{p}_i$ as the mass for each supplier cluster. Our hybrid loss with EMD$^2$-based regularization for classification problems is then defined as:
\begin{equation}
\label{eq:emd-loss-orderless}
\mathrm{E}_H(\mathbf{p}, \mathbf{t}) = \mathrm{E}_X(\mathbf{p}, \mathbf{t}) + \lambda\, \sum\limits_{i=1}^{C}\mathbf{p}_i^2(\mathbf{D}_{i,k}^\omega + \mu) \text{,}
\end{equation} where $\lambda$, $\omega$ and $\mu$ are predefined parameters, such that $\lambda$ defines the weight of EMD$^2$ regularization, the power term $\omega$ determines the sensitivity of the ground distance: a very high $\omega$ means that EMD only penalizes predictions on classes that are far away from the ground truth class, and $\mu$ is the ground distance bias. In our experiments, we use a negative $\mu$ so that $\mathbf{D}_{i,k}^\omega + \mu$ is negative, which means that the network will be rewarded for predictions that are closer to the ground truth class. Note that in Eq. \ref{eq:emd-loss-orderless}, we omitted the $L2$ regularization (weight decay) term which was used in experiments.

\section{Experiments and Results}
\label{sec:experiments}

In this section, we show implementation details and experimental results. First, we demonstrate that our EMD$^2$-based losses outperform the softmax cross-entropy loss on datasets with known strong inter-class relationships: datasets with ordered-classes. At the same time, we show that our self-guided EMD$^2$-based regularization which does not require known inter-class relationships performs as well as the EMD$^2$-based loss that requires strong assumptions on the inter-class relationships. Finally, we show that on datasets without strong inter-class relationships such as ImageNet \cite{russakovsky2015imagenet}, our learned ground distance matrix does not capture spurious inter-class relationships that could lower performance.

\subsection{Implementation details}

We test the EMD$^2$-based losses on different network architectures including the AlexNet \cite{krizhevsky2012imagenet}, VGG 16-layer network \cite{simonyan2014very}, and wide residual network \cite{zagoruyko2016wide}. For optimization, we use stochastic gradient descent with momentum 0.98 in all experiments. The learning rates were selected from $\{10^{-1.5}, 10^{-2}, 10^{-2.5}, 10^{-3}, 10^{-3.5}, 10^{-4}, 10^{-4.5}\}$ individually for each method on each dataset. We notice that when using EMD$^2$ as a regularizer, predicted probability for some classes can be very close to zero, resulting in errors when computing the logarithm of the prediction vector. To solve this, we simply add $1^{-6}$ to the predicted probabilities of all classes. For experiments on ImageNet, we use the same data augmentation and $L2$ weight decay methods used by AlexNet \cite{krizhevsky2012imagenet}. For experiments on all other datasets, we use the following data augmentation methods: during training, we first crop smaller images from the original images (translation augmentation); second, we perturb the images' RGB colors slightly; third, the images are randomly flipped horizontally; fourth, we rotate the images by $(-20, 20)$ degrees; finally, we adjust the aspect ratio by $+/- 10\%$. During testing, we use the average prediction from the center crop and its mirrored image. We use Theano \cite{2016arXiv160502688short} for network implementation.

\subsection{Computational complexity of EMD}
\begin{figure}[t]
\begin{center}
   \includegraphics[trim=4.3cm 8.2cm 4.2cm 7.9cm,clip, width=0.9\linewidth]{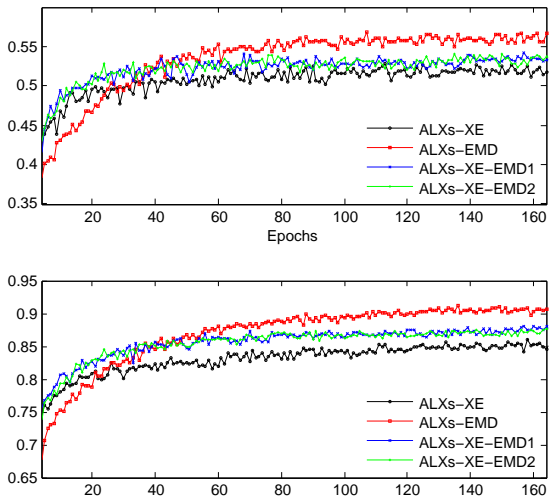}
\end{center}
   \caption{Accuracy of exact match (AEM\%) and with-in-one-category-off match (AEO\%) curves on the Adience dataset (best viewed in color). X-axis: number of training epochs. Y-axis: averaged AEM and AEO results by five-fold cross-validation. Our proposed EMD$^2$-based loss functions outperforms the cross-entropy based loss functions significantly.}
\label{fig:AEM_AEO_curve}
\end{figure}

Our implementation of EMD$^2$-based loss functions adds less than 10\% of CNNs' training time for each iteration and no additional test time. The increment of training time is due to two introduced procedures. However, both of them add very marginal computational complexity. First, the computation time of the loss function is very limited compared to the training time of the entire CNN. Second, to compute the ground distance matrix, no additional CNN forwarding process is needed: we simply store the features of each instance during each training iteration. Moreover, as shown in Fig. \ref{fig:AEM_AEO_curve}, we find that CNNs with the EMD$^2$-based losses achieves the same performance as CNNs with the cross-entropy loss with only $1/3$ of its training epochs.

\subsection{Methods tested}

We describe networks we use below:
\begin{description}
\setlength\itemsep{-0.2em}
\item[ALX] We train the AlexNet (ALX) \cite{krizhevsky2012imagenet} from scratch, to compare with the published baselines \cite{levi2015age,kong2016photo}. For experiments on the Adience dataset, we test a smaller version of AlexNet following the baseline method \cite{levi2015age}. We name it as \textbf{ALXs}.
\item[VGG\textsubscript{F}] We fine-tune the VGG 16-layer network (VGG) \cite{simonyan2014very} pre-trained on ImageNet, to compare with the published baseline \cite{rothe2016deep}.
\item[RES] We use a 40-layer residual network with identity mapping and bottleneck design that is the same as \cite{zagoruyko2016wide}. We train this network from scratch.
\item[RES\textsubscript{F}] We fine-tune a RES pre-trained on ImageNet.
\end{description}

The tested loss functions are described below:
\begin{description}
\setlength\itemsep{-0.2em}
\item[XE] The softmax cross-entropy loss.

\item[REG] The $L2$ regression loss. To use this loss function, the output neurons of a regression network has linear activation functions, instead of softmax, following the conventional regression CNN approach \cite{belagiannis2015robust}. Other parts of the regression network is identical to a classification network.

\item[EMD] The EMD$^2$ loss defined by Eq. \ref{eq:emd-loss-ordered} on ordered-class classification problems.

\item[XEMD1] The self-guided EMD$^2$ regularization training defined by Eq. \ref{eq:emd-loss-orderless}. For the predefined parameters in Eq. \ref{eq:emd-loss-orderless}, we choose $\omega=1$ and $\mu=0.5$. We ``jump-started'' the networks by training with softmax cross-entropy for the first 4 epochs with $\lambda=0$, as a way to avoid using inaccurately estimated ground distance matrix $\mathbf{D}$ for computing class representations. After 4 epochs we choose a $\lambda$ such that the EMD$^2$ term is 3 to 4 times smaller than the cross-entropy term. We find that in ordered-class datasets, the performance was not sensitive when we changed $\lambda$.

\item[XEMD2] The self-guided EMD$^2$ regularization training defined by Eq. \ref{eq:emd-loss-orderless}, but with $\omega=2$ and $\mu=0.25$ and keep other parameters unchanged.

\item[A-EMD] The approximate EMD loss \cite{frogner2015learning}. It requires a predefined ground distance matrix $\mathbf{D}$. Given a specific network, we used the final estimated $\mathbf{D}$ by XEMD2 with the same network, as the predefined matrix. The number of matrix scaling iterations in \cite{frogner2015learning} is set to 100. The entropic regularizer in \cite{frogner2015learning} is selected from $\{0.1, 1, 10\}$ based on the validation error. We use a Caffe implementation of this loss function \cite{jia2014caffe}.
\end{description}

We test different loss functions on various networks. For example, \textbf{ALX-XE} is the AlexNet with the softmax cross-entropy loss.

\begin{table}[t]
\centering
\begin{tabular}{rlllrll}
\hline
\textbf{ALXs\Large{-}} & AEM & AEO & & \textbf{VGG\textsubscript{F}\Large{-}} & AEM & AEO \\
\hline
XE & 53.0 & 85.7 & & XE & 60.9 & 92.8 \\
REG & 49.8 & 85.1 & & REG & 56.5 & \textbf{94.0} \\
EMD & \textbf{57.0} & \textbf{90.4} & & EMD & 59.2 & 92.6 \\
XEMD1 & 54.2 & 88.0 & & XEMD1 & 60.7 & 93.7 \\
XEMD2 & 54.8 & 87.5 & & XEMD2 & \textbf{61.1} & \textbf{94.0} \\
A-EMD & 53.9 & 88.7 & & A-EMD & 60.1 & 93.4 \\
\hline
\hline
\textbf{RES\Large{-}} & AEM & AEO & & \textbf{RES\textsubscript{F}\Large{-}} & AEM & AEO \\
\hline
XE    & 58.1 & 90.3 & & XE    & 60.1 & 92.1 \\
REG   & 57.3 & 91.8 & & REG   & 60.8 & \textbf{94.3} \\
EMD   & \textbf{61.9} & \textbf{93.1} & & EMD   & \textbf{62.2} & \textbf{94.3} \\
XEMD1 & 60.2 & 92.4 & & XEMD1 & 61.9 & 93.8 \\
XEMD2 & 60.2 & 92.7 & & XEMD2 & 61.5 & 94.2 \\
A-EMD & 58.7 & 91.7 & & A-EMD & 59.6 & 92.5 \\
\hline
\hline
\multicolumn{5}{r}{} & AEM & AEO \\
\hline
\multicolumn{5}{r}{Dropout-SVM \cite{eidinger2014age}} & 45.1 & 79.5 \\
\multicolumn{5}{r}{ALXs-XE by Levi \cite{levi2015age}} & 50.7 & 84.7 \\
\multicolumn{5}{r}{Cascade CNN \cite{chencascaded}} & 52.9 & 88.5 \\
\multicolumn{5}{r}{VGG\textsubscript{F}-DEX \cite{rothe2016deep}} & 55.6 & 89.7 \\
\multicolumn{5}{r}{VGG\textsubscript{F}-DEX + IMDB-WIKI \cite{rothe2016deep}} & \textbf{64.0} & \textbf{96.6} \\
\hline
\end{tabular}
\vspace{0.2cm}
\caption{Accuracy of exact match (AEM\%) and with-in-one-category-off match (AEO\%) results on the Adience dataset \cite{eidinger2014age}. The pre-trained VGG network fine-tuned using our proposed loss function (VGG\textsubscript{F}-XEMD2) outperforms the state-of-the-art Deep Expectation (VGG\textsubscript{F}-DEX) method \cite{rothe2016deep} on the same dataset. Our self-guided methods (XEMD1, XEMD2) perform as well as the method with prior knowledge (EMD). The VGG\textsubscript{F}-DEX + IMDB-WIKI \cite{rothe2016deep} method achieves better results using an external age estimation dataset IMDB-WIKI for training, which is 10 times larger than Adience.}
\label{tab:adience-results}
\end{table}

\begin{table*}[t]
\centering
\begin{tabular}{c|cccccccc}
\hline
Age-Groups & 0-2 & 4-6 & 8-13 & 15-20 & 25-32 & 38-43 & 48-53 & 60-\\
\hline
0-2   &    0  &  0.31  &  0.67  &  0.83  &  0.79  &  0.88  &  0.93  &  1.00 \\
4-6   & 0.31  &     0  &  0.31  &  0.63  &  0.69  &  0.81  &  0.87  &  0.93 \\
8-13  & 0.67  &  0.31  &     0  &  0.31  &  0.50  &  0.69  &  0.74  &  0.88 \\
15-20 & 0.83  &  0.63  &  0.31  &     0  &  0.24  &  0.50  &  0.62  &  0.76 \\
25-32 & 0.79  &  0.69  &  0.50  &  0.24  &     0  &  0.31  &  0.56  &  0.71 \\
38-43 & 0.88  &  0.81  &  0.69  &  0.50  &  0.31  &     0  &  0.37  &  0.51 \\
48-53 & 0.93  &  0.87  &  0.74  &  0.62  &  0.56  &  0.37  &     0  &  0.24 \\
60-   & 1.00  &  0.93  &  0.88  &  0.76  &  0.71  &  0.51  &  0.24  &     0 \\
\hline
\end{tabular}
\vspace{0.2cm}
\caption{The learned ground distance matrix $\mathbf{D}$ (Eq. \ref{eq:ground_distance_calculation}) by RES\textsubscript{F}-XEMD2 on the Adience dataset. The age-groups that are further away from each other always have larger ground distances than age-groups that are closer to each other. The interesting semantic differences are also captured during training (e.g. people in their 15-20s are more similar to 25-32s than to 8-13s). }
\label{tab:ground_distance_adience}
\end{table*}

\subsection{Age estimation on Adience dataset}
Age estimation using human face images is important for analyzing and understanding human faces \cite{ricanek2006morph,ramanathan2006face,park2010age}. We test our method on the Adience dataset \cite{eidinger2014age}. The Adience dataset contains 26,000 images in 8 age-groups, and a five-fold cross-validation evaluation scheme. For comparison, we use a smaller version of AlexNet \cite{levi2015age} and fine-tune a pre-trained VGG 16-layer network \cite{rothe2016deep} as described in the earlier experimental details. We use the conventional accuracy of exact match (AEM\%) and with-in-one-category-off match (AEO\%) as evaluation metrics.

The results are shown in Tab. \ref{tab:adience-results}. The pre-trained VGG network fine-tuned using our proposed EMD$^2$-based loss function (VGG\textsubscript{F}-XEMD2) outperforms the state-of-the-art Deep Expectation (VGG\textsubscript{F}-DEX) method \cite{rothe2016deep} on the same dataset. Our self-guided methods (XEMD1, XEMD2) perform as well as the method with prior knowledge (EMD). The VGG\textsubscript{F}-DEX + IMDB-WIKI \cite{rothe2016deep} method achieves better results using an external age estimation dataset IMDB-WIKI which is 10 times larger than Adience. Therefore, our method improves the state-of-the-art when training without external face datasets. The $L2$ regression loss based methods have low AEMs. We believe the reason is that the $L2$ loss is sensitive to outliers. We show the AEM and AEO results with respect to the number of training epochs in Fig. \ref{fig:AEM_AEO_curve}, and the learned ground distance matrix $\mathbf{D}$ (Eq. \ref{eq:ground_distance_calculation}) by our best performing method RES\textsubscript{F}-XEMD2 on the Adience dataset in Tab. \ref{tab:ground_distance_adience}. We can see that age-groups that are further away from each other always have larger ground distances than age-groups that are closer to each other. In addition, the ground distance matrix reflects similarities between different age-groups, \eg, people in their 15-20s are more similar to 25-32s than to 8-13s.

\subsection{Age estimation on Images of Groups dataset}
We further test our method on the Images of Groups dataset \cite{gallagher2009understanding}. This dataset contains 3,500 training face images and 1,000 testing face images in 7 age-groups. We train wide residual networks \cite{zagoruyko2016wide} from scratch and fine-tun a pre-trained VGG 16-layer network \cite{rothe2016deep} on this dataset. The results are shown in Tab. \ref{tab:images-of-groups-results}. Our EMD$^2$-based losses outperforms the cross-entropy loss and the $L2$ loss (regression) in a majority of the cases, with our results achieving a new state-of-the-art on this dataset. Our self-guided methods (XEMD1, XEMD2) perform as well as the method with prior knowledge (EMD).

\begin{table}[t]
\centering
\begin{tabular}{rlllrll}
\hline
\textbf{RES\Large{-}} & AEM & AEO & & \textbf{RES\textsubscript{F}\Large{-}} & AEM & AEO \\
\hline
XE    & 60.0 & 91.5 & & XE    & 61.8 & 94.2 \\
REG   & 52.8 & 92.2 & & REG   & 61.4 & 95.7 \\
EMD   & 59.3 & 92.5 & & EMD   & 63.1 & 95.3 \\
XEMD1 & 58.2 & 91.7 & & XEMD1 & 64.2 & \textbf{96.1} \\
XEMD2 & \textbf{60.1} & \textbf{93.1} & & XEMD2 & \textbf{64.5} & 95.8 \\
\hline
\hline
\textbf{VGG\textsubscript{F}\Large{-}} & AEM & AEO & & & AEM & AEO \\
\hline
XE    & 64.3 & 95.6 & \multicolumn{2}{r}{Single-} & \multirow{2}{*}{54} & \multirow{2}{*}{90} \\
REG   & 60.2 & \textbf{96.6} & \multicolumn{2}{r}{CNN \cite{dong2016automatic}} &  &  \\
EMD   & \textbf{65.0} & 96.1 & \multicolumn{2}{r}{Multi-} & \multirow{2}{*}{56} & \multirow{2}{*}{92} \\
XEMD1 & 63.8 & 95.4 & \multicolumn{2}{r}{CNN \cite{dong2016automatic}} &  &  \\
XEMD2 & 64.6 & 96.1 & &  &  &  \\
\hline
\end{tabular}
\vspace{0.2cm}
\caption{Accuracy of exact match (AEM\%) and with-in-one-category-off match (AEO\%) results on the Images of Groups dataset \cite{gallagher2009understanding}. Our EMD$^2$-based losses outperforms the cross-entropy loss and the $L2$ loss (regression) based methods in majority of the cases. Our self-guided methods (XEMD1, XEMD2) perform as well as the method with prior knowledge (EMD). We improve the state-of-the-art on this dataset.}
\label{tab:images-of-groups-results}
\end{table}

\subsection{Image aesthetics}

\begin{table}[t]
\centering
\begin{tabular}{rllrl}
\hline
\textbf{RES\Large{-}} & $\rho$ & & \textbf{RES\textsubscript{F}\Large{-}} & $\rho$ \\
\hline
XE    & 0.5003 & & XE    &  0.6693 \\
REG   & 0.5235 & & REG   &  0.6609 \\
EMD   & \textbf{0.5448} & & EMD   & \textbf{0.6768} \\
XEMD1 & 0.5370 & & XEMD1 & 0.6751 \\
XEMD2 & 0.5147 & & XEMD2 & 0.6756 \\
\hline
\hline
\textbf{VGG\textsubscript{F}\Large{-}} & $\rho$ & & & $\rho$ \\
\hline
XE    & 0.6283 & & ALX-XE by \cite{kong2016photo} &  0.5923   \\
REG   & 0.6096 & & *Best of \cite{kong2016photo} & 0.6782  \\
EMD   & \textbf{0.6682} & & VGG\textsubscript{F}-XEMD $\times$ 8 & \textbf{0.6889 }  \\
XEMD1 & 0.6371 & & &   \\
XEMD2 & 0.6297 & & &   \\
\hline
\end{tabular}
\vspace{0.2cm}
\caption{Spearmans' $\rho$ results on the image aesthetics with attributes database (AADB) \cite{kong2016photo}. *: used additional 11 labels of image attributes such as color harmony, and image content information. Our EMD$^2$-based losses outperforms cross-entropy loss and $L2$ loss (regression) based methods significantly. By averaging the results of eight VGG\textsubscript{F}-EMD networks, the results with just image data are better than the previous state-of-the-art, various attribute-augmented model \cite{kong2016photo}.}
\label{tab:image-aesthetics-results}
\end{table}

Assessing image aesthetics automatically has a wide range of applications \cite{lu2014rapid,datta2006studying,datta2008image}. We test our method on the Image Aesthetics with Attributes Database (AADB) \cite{kong2016photo} which contains 8,458 training and 1,000 testing images, labeled as real numbers in $[0.0, 1.0]$ according to the viewers' aesthetic judgments. To transform this dataset into a classification dataset, we discretize the real number labels to 10 bins, balancing the number of training images in each bin. During testing, we compute the expected aesthetic scores according to the predicted distributions of 10 aesthetic bins. This give us real-numbered predictions. We use Spearmans' rank correlation $\rho$ as the evaluation metric, following \cite{kong2016photo}.

\begin{table}[t]
\centering
\begin{tabular}{lll}
\hline
Method & Top-1 Error & Top-5 Error  \\
\hline
\hline
ALX-XE & 0.435 & 0.207 \\
ALX-XEMD1 & \textbf{0.422} & \textbf{0.202} \\
RES-XE & \textbf{0.232} & 0.0657 \\
RES-XEMD1 & 0.233 & \textbf{0.0652} \\
\hline
\end{tabular}
\vspace{0.2cm}
\caption{Experimental results on the ImageNet ILSVRC 2012 dataset \cite{russakovsky2015imagenet}. Our self-guided method can be applied on general datasets with no adverse performance effects. Our method finds that there is no strong inter-class relationships on this dataset (details in text).}
\label{tab:imagenet-results}
\end{table}

The results are shown in Tab. \ref{tab:image-aesthetics-results}. Our EMD$^2$-based losses again outperform cross-entropy loss and $L2$ loss (regression) significantly. We conduct additional experiments by discretizing the real-numbered aesthetic labels to 8 different number of bins (3,4,5,6,7,8,9,10 bins), which give us 8 sets of ground truth labels. Then, we fine-tune one VGG\textsubscript{F}-EMD network for each set of ground truth and average the prediction results into an ensemble model, and denote this method as VGG\textsubscript{F}-EMD $\times$ 8. It achieves state-of-the-art results with only image training data, outperforming the previous state-of-the-art method trained with additional 11 labels such as color harmony and vivid color information.

\subsection{Generalization on ImageNet}
We show the generalization ability of our self-guided EMD$^2$-based regularizer (Eq. \ref{eq:emd-loss-orderless}) on the ImageNet ILSVRC 2012 dataset \cite{russakovsky2015imagenet}, which is a classification dataset with weak inter-class relationships. We test the original AlexNet and 40-layer residual network on this dataset with cross-entropy, and separately with the self-guided EMD$^2$ regularization. We do not test the VGG network because training it from scratch is time consuming. The results of the validations set are reported in Tab. \ref{tab:imagenet-results}. We see that one can safely apply our self-guided method on datasets with weak inter-class relationships.

Our method does not outperform the baseline because our self-guided method finds that the inter-class relationship on this dataset is less significant compared to those in ordered-class datasets. We show this by measuring the Standard Deviation of pair-wise Distances between the estimated centroids of the classes (SDD, standard deviation between all entries of $\mathbf{\bar D}$ introduced in Sec. \ref{sec:estimating_gd_matrix}). A higher SDD indicates a stronger inter-class relationship. On the Adience dataset, SDD$=0.0335$; On Images of Groups SDD$=0.0164$; On AADB, SDD$=0.0184$; On ImageNet, SDD$=0.00614$.

\section{Conclusion}
\label{sec:conclusion}

In this work, we argued that the conventional softmax cross-entropy loss for training CNNs only maximizes the predicted probability at the ground truth label, and ignores the inter-class relationships. We proposed to use the exact squared earth mover's distance (EMD$^2$) in loss functions for CNN training to take class relationships into account. We evaluated our methods on two age estimation datasets and one image aesthetic assessment dataset. Our method significantly outperformed state-of-the-art regression-based and cross-entropy-based CNNs using no external datasets, and with only image information. Furthermore, we demonstrated that our method can discover the inter-class relationships efficiently with no prior knowledge. Finally, we showed that our method can be applied to datasets with weak inter-class relationships with no adverse results. Our future works include a more sophisticated ground distance matrix computing method, and exploring variants of EMDs that can perform better on datasets with weak inter-class relationships.

{\small
\bibliographystyle{ieee}
\bibliography{egbib}

\begin{thebibliography}{10}\itemsep=-1pt

\bibitem{arjovsky2017wasserstein}
M.~Arjovsky, S.~Chintala, and L.~Bottou.
\newblock Wasserstein gan.
\newblock {\em arXiv preprint arXiv:1701.07875}, 2017.

\bibitem{belagiannis2015robust}
V.~Belagiannis, C.~Rupprecht, G.~Carneiro, and N.~Navab.
\newblock Robust optimization for deep regression.
\newblock In {\em ICCV}, 2015.

\bibitem{bogachev2012monge}
V.~I. Bogachev and A.~V. Kolesnikov.
\newblock The monge-kantorovich problem: achievements, connections, and
  perspectives.
\newblock {\em Russian Mathematical Surveys}, 67, 2012.

\bibitem{chencascaded}
J.-C. Chen, A.~Kumar, R.~Ranjan, V.~M. Patel, A.~Alavi, and R.~Chellappa.
\newblock A cascaded convolutional neural network for age estimation of
  unconstrained faces.
\newblock In {\em Biometrics Theory, Applications and Systems (BTAS)}, 2016.

\bibitem{datta2006studying}
R.~Datta, D.~Joshi, J.~Li, and J.~Z. Wang.
\newblock Studying aesthetics in photographic images using a computational
  approach.
\newblock In {\em ECCV}, 2006.

\bibitem{datta2008image}
R.~Datta, D.~Joshi, J.~Li, and J.~Z. Wang.
\newblock Image retrieval: Ideas, influences, and trends of the new age.
\newblock {\em ACM Computing Surveys (CSUR)}, 40, 2008.

\bibitem{deng2014ensemble}
L.~Deng and J.~C. Platt.
\newblock Ensemble deep learning for speech recognition.
\newblock In {\em Fifteenth Annual Conference of the International Speech
  Communication Association}, 2014.

\bibitem{dong2014learning}
C.~Dong, C.~C. Loy, K.~He, and X.~Tang.
\newblock Learning a deep convolutional network for image super-resolution.
\newblock In {\em ECCV}, 2014.

\bibitem{dong2016automatic}
Y.~Dong, Y.~Liu, and S.~Lian.
\newblock Automatic age estimation based on deep learning algorithm.
\newblock {\em Neurocomputing}, 187, 2016.

\bibitem{eidinger2014age}
E.~Eidinger, R.~Enbar, and T.~Hassner.
\newblock Age and gender estimation of unfiltered faces.
\newblock {\em IEEE Transactions on Information Forensics and Security}, 2014.

\bibitem{escalera2015chalearn}
S.~Escalera, J.~Fabian, P.~Pardo, X.~Bar{\'o}, J.~Gonzalez, H.~J. Escalante,
  D.~Misevic, U.~Steiner, and I.~Guyon.
\newblock Chalearn looking at people 2015: Apparent age and cultural event
  recognition datasets and results.
\newblock In {\em ICCV Workshop}, 2015.

\bibitem{frogner2015learning}
C.~Frogner, C.~Zhang, H.~Mobahi, M.~Araya, and T.~A. Poggio.
\newblock Learning with a wasserstein loss.
\newblock In {\em NIPS}, 2015.

\bibitem{gallagher2009understanding}
A.~C. Gallagher and T.~Chen.
\newblock Understanding images of groups of people.
\newblock In {\em CVPR}, 2009.

\bibitem{girshick2014rich}
R.~Girshick, J.~Donahue, T.~Darrell, and J.~Malik.
\newblock Rich feature hierarchies for accurate object detection and semantic
  segmentation.
\newblock In {\em CVPR}, 2014.

\bibitem{golik2013cross}
P.~Golik, P.~Doetsch, and H.~Ney.
\newblock Cross-entropy vs. squared error training: a theoretical and
  experimental comparison.
\newblock In {\em Interspeech}, 2013.

\bibitem{hannun2014deep}
A.~Hannun, C.~Case, J.~Casper, B.~Catanzaro, G.~Diamos, E.~Elsen, R.~Prenger,
  S.~Satheesh, S.~Sengupta, A.~Coates, et~al.
\newblock Deep speech: Scaling up end-to-end speech recognition.
\newblock {\em arXiv preprint arXiv:1412.5567}, 2014.

\bibitem{he2015deep}
K.~He, X.~Zhang, S.~Ren, and J.~Sun.
\newblock Deep residual learning for image recognition.
\newblock {\em CVPR}, 2015.

\bibitem{jia2014caffe}
Y.~Jia, E.~Shelhamer, J.~Donahue, S.~Karayev, J.~Long, R.~Girshick,
  S.~Guadarrama, and T.~Darrell.
\newblock Caffe: Convolutional architecture for fast feature embedding.
\newblock In {\em ACM international conference on Multimedia}.

\bibitem{kong2016photo}
S.~Kong, X.~Shen, Z.~Lin, R.~Mech, and C.~Fowlkes.
\newblock Photo aesthetics ranking network with attributes and content
  adaptation.
\newblock {\em arXiv preprint arXiv:1606.01621}, 2016.

\bibitem{krizhevsky2012imagenet}
A.~Krizhevsky, I.~Sutskever, and G.~E. Hinton.
\newblock Imagenet classification with deep convolutional neural networks.
\newblock In {\em NIPS}, 2012.

\bibitem{lecun1998gradient}
Y.~LeCun, L.~Bottou, Y.~Bengio, and P.~Haffner.
\newblock Gradient-based learning applied to document recognition.
\newblock {\em Proceedings of the IEEE}, 86(11), 1998.

\bibitem{levi2015age}
G.~Levi and T.~Hassner.
\newblock Age and gender classification using convolutional neural networks.
\newblock In {\em CVPR Workshop}, 2015.

\bibitem{levina2001earth}
E.~Levina and P.~Bickel.
\newblock The earth mover's distance is the mallows distance: some insights
  from statistics.
\newblock In {\em Computer Vision, 2001. ICCV 2001. Proceedings. Eighth IEEE
  International Conference on}, volume~2, 2001.

\bibitem{li2015convolutional}
H.~Li, Z.~Lin, X.~Shen, J.~Brandt, and G.~Hua.
\newblock A convolutional neural network cascade for face detection.
\newblock In {\em Proceedings of the IEEE Conference on Computer Vision and
  Pattern Recognition}, 2015.

\bibitem{liu2014recursive}
S.~Liu, N.~Yang, M.~Li, and M.~Zhou.
\newblock A recursive recurrent neural network for statistical machine
  translation.
\newblock In {\em ACL (1)}, 2014.

\bibitem{lu2014rapid}
X.~Lu, Z.~Lin, H.~Jin, J.~Yang, and J.~Z. Wang.
\newblock Rapid: rating pictorial aesthetics using deep learning.
\newblock In {\em ACM international conference on Multimedia}, 2014.

\bibitem{luenberger1973introduction}
D.~G. Luenberger.
\newblock {\em Introduction to linear and nonlinear programming}, volume~28.
\newblock 1973.

\bibitem{ma2016texture}
K.~Ma, D.~Samaras, M.~Petrucci, D.~L. Magnus, et~al.
\newblock Texture classification for rail surface condition evaluation.
\newblock In {\em Winter Conference on Applications of Computer Vision (WACV)},
  2016.

\bibitem{martinez2016relaxed}
M.~Martinez, M.~Haurilet, Z.~Al-Halah, M.~Tapaswi, and R.~Stiefelhagen.
\newblock Relaxed earth mover's distances for chain-and tree-connected spaces
  and their use as a loss function in deep learning.
\newblock {\em arXiv preprint arXiv:1611.07573}, 2016.

\bibitem{park2010age}
U.~Park, Y.~Tong, and A.~K. Jain.
\newblock Age-invariant face recognition.
\newblock {\em IEEE transactions on pattern analysis and machine intelligence},
  32, 2010.

\bibitem{ramanathan2006face}
N.~Ramanathan and R.~Chellappa.
\newblock Face verification across age progression.
\newblock {\em IEEE Transactions on Image Processing}, 15, 2006.

\bibitem{ricanek2006morph}
K.~Ricanek and T.~Tesafaye.
\newblock Morph: A longitudinal image database of normal adult age-progression.
\newblock In {\em International Conference on Automatic Face and Gesture
  Recognition (FGR06)}, 2006.

\bibitem{rothe2016deep}
R.~Rothe, R.~Timofte, and L.~Van~Gool.
\newblock Deep expectation of real and apparent age from a single image without
  facial landmarks.
\newblock {\em IJCV}, 2016.

\bibitem{rubner2000earth}
Y.~Rubner, C.~Tomasi, and L.~J. Guibas.
\newblock The earth mover's distance as a metric for image retrieval.
\newblock {\em IJCV}, 40, 2000.

\bibitem{russakovsky2015imagenet}
O.~Russakovsky, J.~Deng, H.~Su, J.~Krause, S.~Satheesh, S.~Ma, Z.~Huang,
  A.~Karpathy, A.~Khosla, M.~Bernstein, et~al.
\newblock Imagenet large scale visual recognition challenge.
\newblock {\em IJCV}, 115, 2015.

\bibitem{shalev2011stochastic}
S.~Shalev-Shwartz and A.~Tewari.
\newblock Stochastic methods for l1-regularized loss minimization.
\newblock {\em Journal of Machine Learning Research}, 12, 2011.

\bibitem{simonyan2014very}
K.~Simonyan and A.~Zisserman.
\newblock Very deep convolutional networks for large-scale image recognition.
\newblock {\em ICLR}, 2014.

\bibitem{socher2011parsing}
R.~Socher, C.~C. Lin, C.~Manning, and A.~Y. Ng.
\newblock Parsing natural scenes and natural language with recursive neural
  networks.
\newblock In {\em ICML}, 2011.

\bibitem{sutic2010automatic}
D.~Suti{\'c}, I.~Bre{\v{s}}kovi{\'c}, R.~Hui{\'c}, and I.~Juki{\'c}.
\newblock Automatic evaluation of facial attractiveness.
\newblock In {\em MIPRO, 2010 Proceedings of the 33rd International
  Convention}, 2010.

\bibitem{2016arXiv160502688short}
{Theano Development Team}.
\newblock {Theano: A {Python} framework for fast computation of mathematical
  expressions}.
\newblock {\em arXiv preprint arXiv:1605.02688}, 2016.

\bibitem{zagoruyko2016wide}
S.~Zagoruyko and N.~Komodakis.
\newblock Wide residual networks.
\newblock {\em BMVC}, 2016.

\end{thebibliography}
}

\end{document}